\documentclass[10pt,a4paper]{article}
\usepackage[utf8]{inputenc}
\usepackage{amsmath}
\usepackage{amsfonts}
\usepackage{amssymb}
\usepackage{graphicx}
\usepackage[left=2cm,right=2cm,top=2cm,bottom=2cm]{geometry}

\usepackage[numbers]{natbib}
\usepackage{authblk}
\usepackage{latexsym}
\usepackage{algorithm}
\usepackage{algorithmic}
\usepackage{placeins}

\usepackage{url}
\usepackage{xcolor}

\DeclareMathOperator*{\argmin}{arg\,min}

\begin{document}

\title{Random Forest regression for manifold-valued responses}

\author{Dimosthenis Tsagkrasoulis}
\affil{Department of Mathematics, Imperial College London, SW7 2AZ, London, UK}
\author{Giovanni Montana
\thanks{Corresponding author:  giovanni.montana@kcl.ac.uk}
}
\affil{Department of Biomedical Engineering, King's College London, SE1 7EH, London, UK}
\affil{Department of Mathematics, Imperial College London, SW7 2AZ, London, UK}

\date{}
\maketitle

\begin{abstract}
An increasing array of biomedical and computer vision applications requires the predictive modeling of complex data, for example images and shapes. The main challenge when predicting such objects lies in the fact that they do not comply to the assumptions of Euclidean geometry. Rather, they occupy non-linear spaces, a.k.a. manifolds, where it is difficult to define concepts such as coordinates, vectors and expected values. In this work, we construct a non-parametric predictive methodology for manifold-valued objects, based on a distance modification of the Random Forest algorithm. Our method is versatile and can be applied both in cases where the response space is a well-defined manifold, but also when such knowledge is not available. Model fitting and prediction phases only require the definition of a suitable distance function for the observed responses. We validate our methodology using simulations and apply it on a series of illustrative image completion applications, showcasing superior predictive performance, compared to various established regression methods.
\end{abstract}

\section{Introduction} \label{intro}

Predictive modeling is an integral part of data analysis. It encompasses the process of developing models which can accurately predict yet-to-be-seen data. A multitude of regression models exist that can be used for prediction of univariate, or multivariate vectorial, responses. Nevertheless, less work has been done on the difficult problem of modeling and predicting more complex objects that possess additional structure, be it morphological, directional, or otherwise \cite{Rahman2005}.

Images \cite{peyre2009}, shapes \cite{Small1996}, graphs \cite{Tsochantaridis2004}, deformation tensors \cite{Zhu2009} are examples of complex data types that appear naturally as responses in image analysis, computer vision, medical imaging and other application areas. While such objects are typically represented as points on very high-dimensional spaces, they can be meaningfully represented as points lying on smooth non-linear hyper-surfaces of lower dimensionality, a.k.a manifolds \cite{Chavel2006}. Manifolds can be understood as surfaces that locally resemble the Euclidean plane, but have different global structures \cite{robbin2011}. Ideally, a suitable predictive modeling methodology should work under the additional constrains imposed by the data's inherent manifold structure instead of simply treating these complex objects as points on Euclidean spaces. Major difficulties of course arise by forgoing the assumption of a Euclidean space, such as lack of coordinates, vectors and no analytical definitions of expected values \cite{Fletcher2013}. In this work, we are additionally interested in experimental settings where the input observations may be very-high dimensional, which poses further requirements in the construction of a predictive algorithm.

Three main families of methodologies concerning with regression for manifold-valued responses can be found in the literature; intrinsic manifold regression, kernel-based and Manifold Learning (ML)-based methods. Intrinsic manifold regression models are generalizations of linear regression on manifolds \cite{Fletcher2013, Zhu2009}. They require the analytical definition of the data manifold, since they fit a parametrized curve on the data.The assumption of an underlying manifold drives the choice of geometric elements employed during model definition and parameter estimation. Unfortunately, this requirement can be rarely satisfied, either due to the nature of data or the inability to define a suitable manifold. Furthermore, this methodology is not ideal for regression analysis with highly dimensional input observations.

Kernel methods first appeared in the literature as methodologies tailored for complex input objects, such as trees or graphs \cite{shawe2004}. The basic idea behind these methods is that, if the input data lie on a non-linear space, then they can be implicitly mapped on a very high (or infinite) dimensional inner-product space, in which standard regression methods can be applied \cite{shawe2004}. This implicit mapping is achieved through a kernel function that defines the objects' inner-product in that space \cite{shawe2004}. Kernel methods for complex responses have also been proposed \cite{Tsochantaridis2004, Geurts2006}, but suffer from a number of issues. First, depending on the data at hand, a meaningful kernel function must be found or constructed, a process that is not always intuitive. Second, another problem is the formulation of a prediction methodology, which has to be described as a kernel minimization problem over the response space. This is most commonly solved by reducing the search space to the training dataset.

The last family of manifold regression methods is based on non-linear dimensionality reduction, a.k.a Manifold Learning. Given a set of observed points, ML methods aim to project these onto a space of lower dimensionality, whilst retaining as well as possible the original geometrical structure of the data. ML-based methods first apply ML on the response data, and subsequently use standard regression models trained on the learned response embedding. A suitable methodology must be formulated to map predicted points from the embedding to the manifold, a process that is referred to as backscoring \cite{wahba1992}. Appropriate selection of the ML technique is largely based on intuition and previous experience. A decrease in model fitting accuracy, compared to an intrinsic manifold model, is to be expected, since the response embedding is not guaranteed to completely capture the structure of the underlying manifold. Furthermore, ML techniques generate discrete and one way maps from the manifold to the embedding, and the existence of an inverse and continuous map cannot always be guaranteed \cite{wahba1992}.

Here, we present a new methodology for predictive modeling of manifold-valued responses, which addresses a number of key issues listed above. Our objective was to propose a unified framework for regression and prediction of complex objects that is accurate, computationally efficient and can be readily deployed in a variety of applications. In the training phase, we employ our modified Random Forest (RF) regression algorithm that can be trained using only pairwise distances between response observations \cite{Sim2013}. The non-parametric RF methodology is efficient and can handle high dimensional input spaces. In contrast to intrinsic manifold regression, no analytical definition of the underlying response manifold is required, apart from the construction of an appropriate distance function for the responses. Comparing our distance-based approach to kernel methods, we identify further advantages. First, a vast library of readily available distance metrics exists in the literature (see for example \cite{deza2009}). Second, when the manifold metric is not known, distances can be intuitively approximated, using for example the Isomap distance formulation \cite{Tenenbaum2000}.

In the prediction phase, a response point estimate for a new input observation is found as follows. Pairwise distances between the unseen response and all training observations are estimated. Using this set of distances, the response is predicted on a Euclidean embedding computed using the ML technique of metric multi-dimensional scaling (MDS) and is finally mapped to the response manifold through a standardized backscoring procedure. Our prediction methodology follows a two-step approach akin to ML-based methods, whilst offering two additional advantages. First, our regression model is trained on the original manifold, which enhances the quality of the fitted model. Second, due to the fact that manifold distances are known, MDS is employed for ML and predictions can be analytically computed and backscored to the response manifold.

The rest of this article is organized as follows. In section \ref{sec:methods} we present the details of our manifold regression methodology. Our simulation and application experiments are included in section \ref{sec:exp}. We conclude this work in section \ref{sec:disc}.

\section{Random Forest Predictive Methodology for Manifold-Valued Objects} \label{sec:methods}

\subsection{Problem Definition}

\begin{figure}[!ht]
  \centering
  \includegraphics[width=0.6\textwidth]{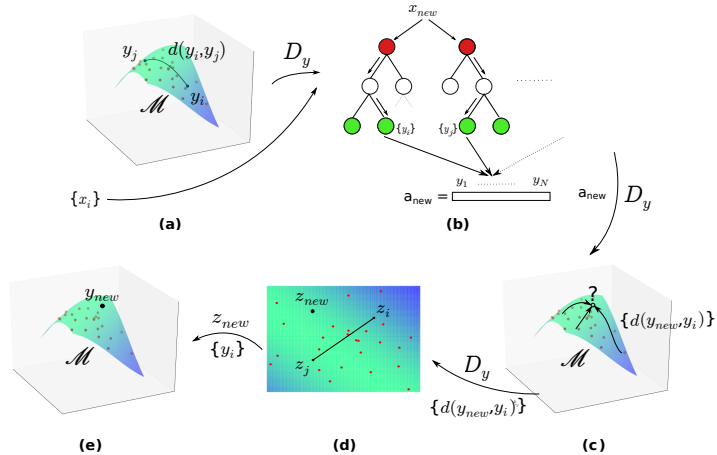}
  \caption {Illustration of the dRF prediction methodology. (a) The dRF model is fitted using the distance matrix $\mathbf{D}$ of manifold distances between the observed responses (section \ref{sec:drf}). (b) When a new input $\mathbf{x}_{new}$ is observed, it is passed through the forest and a similarity vector, $\mathbf{a}$, between the training responses and the yet to be predicted response is extracted from the model (section \ref{sec:dist}). (c) Based on $\mathbf{a}$, the set of distances between the new point and the training responses on the manifold are predicted (section \ref{sec:dist}). (d) Knowledge of these distances allows prediction of the response on a Euclidean embedding of the manifold, extracted through multi-dimensional Scaling (section \ref{sec:outofsample}). (e) A backscoring method is used to project the predicted point back to the original manifold (section \ref{sec:back}).}
\label{man:pred:fig1}
\end{figure}

Let $S$ be a dataset of $N$ observed input-response pairs $\{(\mathbf{x}_i, \mathbf{y}_i)\}_{i=1}^N$, with inputs $\mathbf{x}_i = (x_{i1}, \ldots,x_{ip}) \in \mathbb{R}^p$ corresponding to responses $\mathbf{y}_i \in \mathcal{M} \subset \mathbb{R}^q$. $\mathcal{M}$ is a possibly unknown manifold, equipped with a distance metric $d(\cdot,\cdot)$. We refer to $\mathbb{R}^q$ as the response representation space. Let $\mathbf{D}$ be the $N \times N$ matrix of pairwise distances between the observed response points, with elements $D_{ij} = d(\mathbf{y}_i,\mathbf{y}_j)$. See Fig. \ref{man:pred:fig1}(a) for a graphical illustration of the described data. We want to construct a predictive methodology that leverages these distances in order to ensure that, for any given input $\mathbf{x}_{new}$, the predicted $\mathbf{\hat{y}}_{new}$ lies on $\mathcal{M}$.

\subsection{Distance Random Forest Regression} \label{sec:drf}

We first concentrate on the construction of a suitable manifold regression learning algorithm. Random Forest (RF) is a non-parametric, non-linear regression and classification algorithm \cite{Breiman2001}. In more detail, RF is a collection of Classification and Regression Trees (CARTs). A CART splits the input space recursively, according to a predefined split function, to small rectangular regions and then fits a simple model, commonly a constant value, in each one of them. See \cite{Breiman2001} for a detailed description of RF. In \cite{Sim2013}, we presented a modified distance Random Forest (dRF) algorithm, where the split criterion was formulated to depend only on pairwise distances between responses:
\begin{equation}\label{man:drf:eq:cost_function}
\begin{split}
G_{\mathcal{M}}(S_w) =  \frac{1}{2N_w} \sum_{\mathbf{y}_i \in S_w} \sum_{\mathbf{y}_i \in S_w} d^2 (\mathbf{y}_i,\mathbf{y}_j) - \frac{1}{2N_{wl}} \sum_{\mathbf{y}_i \in S_{wl}} \sum_{\mathbf{y}_i \in S_{wl}} d^2 (\mathbf{y}_i,\mathbf{y}_j) - \frac{1}{2N_{wr}} \sum_{\mathbf{y}_i \in S_{wr}} \sum_{\mathbf{y}_i \in S_{wr}} d^2 (\mathbf{y}_i,\mathbf{y}_j) \enspace ,
\end{split}
\end{equation}
with $\mathbf{y}_i \in \mathcal{M}$. Here $N_w, N_{wl}, N_{wr} $ are the cardinalities of the data subsets $S_w$, $S_{wl}$, $S_{wr}$ belonging to a node $w$ and its children nodes (left and right), respectively. The objective \eqref{man:drf:eq:cost_function} is a generalization of the cost function used in standard regression RF, decoupled from the use of Euclidean norms and means, dependent only on pairwise response distances. As such, it enables the RF algorithm to be applied in general metric spaces. Previously, we used dRF for regression applications with graph and covariance-based response objects \cite{Sim2013}. dRF lends itself naturally to manifold-valued data, whether the manifold in question is analytically defined or implied by the use of a specific distance metric.

\subsection{Predictive Methodology for dRF} \label{sec:pred}

Traditionally, when working with responses lying in Euclidean spaces, an RF prediction in made simply by averaging response points; a new input observation follows the split decision rules learned during training and reaches a terminal node -leaf- in the tree. It is then assigned a suitable value on the response space, most often the average response of that leaf's responses in the training set. This approach though is not valid under the assumption of non-vectorial manifold responses \cite{Pennec2006}. 

Our proposed methodology uses dRF for learning a family of trees from the data, which only requires an appropriate distance metric for the responses, as described above. The prediction phase is substantially different from standard RF. For a new input point $\mathbf{x}_{new}$, we use the trained dRF model to predict all pairwise distances between the unknown response, $\mathbf{y}_{new}$, and all observed responses in the training dataset, $\mathbf{y}_i$. Having estimated these distances, we then utilize them to to predict the response on an Euclidean embedding of the manifold, learned from the observed response set using metric MDS. Subsequently, a backscoring method is employed to project the point back to the original response manifold $\mathcal{M}$. Figure \ref{man:pred:fig1} provides an illustration of the proposed methodology, and the details are in order.

\subsubsection{RF-based Distance Prediction} \label{sec:dist}

The first step is to estimate the set of distances $\{\hat{d}(\mathbf{y}_{new},\mathbf{y}_i)\}_{i=1}^N$. We exploit the inherent ability of the dRF model to provide a measure of similarity between pairs of response observations (see Fig. \ref{man:pred:fig1}(b)). When the new input point $\mathbf{x}_{new}$ is `dropped' through each tree in the forest, it reaches a leaf associated with a subset of the training data. The $\mathbf{y}_{new}$ can be considered similar to that leaf's training responses. To compute the dRF-based similarities, a vector $\mathbf{a} = (a_{1},\ldots,a_{N} )$ is used, with each element $a_i$ corresponding to the similarity between $\mathbf{y}_{new}$ and $\mathbf{y}_i$. Initially $a_i$ is set to zero, for all $i=1, \ldots, N$. For each tree in RF, $a_{i}$ is incremented by one, each time $\mathbf{x}_{new}$ ends in the same node as $\mathbf{x}_{i}$. Normalization of similarities is performed by division with the number of trees.

Based on the similarity vector $\mathbf{a}$ and the training distance matrix $\mathbf{D}$, the response distances are predicted using the algorithmic procedure \ref{man:pred:algo1} ( see Fig. \ref{man:pred:fig1}(c) ), which guarantees that the new point will lie in close proximity to at least its closest neighbor based on the dRF affinities, and that predicted distance values respect the triangular inequality property of metric $d$. In detail, the algorithm initially identifies the closest training point to the new response, according to $\mathbf{a}$, and assigns the minimum distance observed in the training set as the distance between these two responses. Subsequently, we iterate over the remaining responses, in decreasing order of distance from the first point, and assign a value for the distance to the new observation as follows: For each triplet including the considered point, the new point and a point for which the distance to $\mathbf{y}_{new}$ has been already estimated, we keep the maximum of the two known distances. Subsequently, we assign the minimum of all identified maximums as the predicted distance.

\begin{algorithm}
\caption{Prediction of $\{\hat{d}(\mathbf{y}_{new},\mathbf{y}_i)\}_{i=1}^N$ using $\mathbf{a}$ and $\mathbf{D}$.}
\label{man:pred:algo1}
\begin{algorithmic}
\REQUIRE $\mathbf{a}$ \AND $\{ D_{ij} \}_{i,j=1}^N$
\STATE $S = \{1, \ldots ,N \}, \enspace Q = \emptyset$
\STATE $l \gets \underset{i \in S}{\mbox{argmax}} \enspace\ a_i $
\STATE $\hat{d}(\mathbf{y}_{new},\mathbf{y}_{l}) \gets \min \{ D_{ij}\}_{i,j=1}^N$
\STATE $S \gets S \setminus \{ l \}, \enspace Q \gets Q \cup \{ l \}$
\REPEAT
\STATE $p \gets \underset{i \in S}{\mbox{argmax}} \enspace D_{il}  $
\STATE $\hat{d}(\mathbf{y}_{new},\mathbf{y}_{p}) \gets min \{  max \{\hat{d}(\mathbf{y}_{new},\mathbf{y}_{i}), D_{ip} \}| i \in Q  \}  $
\STATE $S \gets S \setminus \{ p \}, \enspace Q \gets Q \cup \{ p \}$
\UNTIL{$S = \emptyset $}
\RETURN $\{\hat{d}(\mathbf{y}_{new},\mathbf{y}_i)\}_{i=1}^N$

\end{algorithmic}

\end{algorithm}

\subsubsection{Prediction on the Response Embedding} \label{sec:outofsample}

Given $\{\hat{d}(\mathbf{y}_{new},\mathbf{y}_i)\}_{i=1}^N$, we can predict a point estimate of the response on a Euclidean embedding of the manifold using MDS \cite{Cox2010} (see Fig. \ref{man:pred:fig1}(d) ).

MDS computes an approximation $\{ \mathbf{z}_i| \mathbf{z}_i \in \mathbb{R}^m \}_{i=1}^N$ of the manifold-valued dataset that resides on a $m$-dimensional Euclidean space, by minimizing a stress function of the form $\sum_{i=1}^N\sum_{j=1}^N \left( d^2(\mathbf{y}_i,\mathbf{y}_j) - \| \mathbf{z}_i - \mathbf{z}_j\|^2 \right)$. 
Let $\mathbf{D}^{(2)}$ be the $N \times N $ matrix of squared manifold distances
and $\mathbf{K} = - \frac{1}{2} \mathbf{H} \mathbf{D}^{(2)} \mathbf{H}$, where $\mathbf{H}=\mathbf{I}_N-\frac{1}{N}\mathbf{e} \mathbf{e}^T$, with $\mathbf{I}_N$ the $N \times N $ identity matrix and $\mathbf{e}$ an $N \times 1$ column vector of all ones.
Individual elements of $\mathbf{K}$ are given by $K_{ij} = - \frac{1}{2} (D^{(2)}_{ij} - \frac{1}{n}S_i - \frac{1}{n}S_j + \frac{1}{n^2}S), i,j = 1, \ldots ,N ,$
where $S_i = \sum_j D^{(2)}_{ij}, S_j = \sum_i D^{(2)}_{ij}, S = \sum_{i,j} D^{(2)}_{ij}$ are the $i^{th}$ row, $j^{th}$ column and overall element-wise sums of $\mathbf{D}^{(2)}$, respectively. 
In \cite{mardia1978}, it was shown that, if $\mathbf{K}$
has rank $p \enspace (p \leq N-1)$, with $\lambda_1, \ldots , \lambda_p$ the $p$ ordered non-zero eigenvalues of $\mathbf{K}$ with corresponding eigenvectors $\mathbf{u}_1, \ldots, \mathbf{u}_p$, and $\mathbf{u}_l = (u_{l1},\ldots, u_{lN})$ for $l = 1, \ldots, p$, then the solution to the MDS problem, 
assuming $m \leq p$, is given by $z_{il} = \sqrt{\lambda_l}u_{li}$, with $l = 1, \ldots, m, \enspace i = 1, \ldots, N$, and $\mathbf{z}_i = (z_{i1},\ldots,z_{im}) $.

The result of the MDS decomposition is a discrete and one-way mapping $\mathbf{z}_i = f_{MDS}(\mathbf{y}_i)$
defined  on the training data set. An Out-Of-Sample (OOS) method which allows mapping of a new manifold observation on the learned space of the embedding was constructed in \cite{Bengio2004}. 
Let $Y \in \mathcal{M}$ be a random variable defined on the manifold surface and $\mathbf{y}_a,\mathbf{y}_b\in \mathcal{M}$ two observations of $Y$. A continuous kernel function $k_{MDS}$, which gives rise to $\mathbf{K}$ under the training observations, is defined as $k_{MDS}(\mathbf{y}_a,\mathbf{y}_b) = - \frac{1}{2} (d^2(\mathbf{y}_a,\mathbf{y}_b) - E[d^2(Y,\mathbf{y}_a)] - E[d^2(Y,\mathbf{y}_b)] + E[d^2(Y,Y)] ) , $
where $E(\cdot)$ denotes expectation. 
Then, an OOS prediction $\hat{\mathbf{z}}_{new} = (\hat{z}_{new,1},\ldots,\hat{z}_{new,m}) $ is given by \cite{Bengio2004}:
\begin{equation}\label{man:pred:eq:oos}
\hat{z}_{new,k} = \frac{1}{\sqrt{\lambda_k}} \sum^N_{i=1} u_{ik} \hat{k}_{MDS}(\mathbf{y}_{new},\mathbf{y}_i), \enspace k = 1,\ldots,m  \enspace ,
\end{equation}
with $\hat{k}_{MDS}$ denoting the mean estimator of $k_{MDS}$ under the augmented dataset $\{ \mathbf{y}_i\}_{i=1}^N \cup \{ \mathbf{y}_{new} \}$:
\begin{equation}
\begin{split}
\hat{k}_{MDS}(\mathbf{y}_{new},&\mathbf{y}_i) = \enspace -\frac{1}{2} d^2(\mathbf{y}_{new},\mathbf{y}_i) + \frac{1}{2(N+1)} \left( \sum_{j=1}^N d^2(\mathbf{y}_j,\mathbf{y}_{new}) \right) \\
& + \frac{1}{2(N+1)} \left( \left( \sum_{j=1}^N d^2(\mathbf{y}_j,\mathbf{y}_i) \right) + d^2(\mathbf{y}_{new},\mathbf{y}_i) \right) \\
& - \frac{1}{2(N+1)^2} \left( \left( \sum_{j,l=1}^N d^2(\mathbf{y}_{j},\mathbf{y}_l) \right) + 2\sum_{j=1}^N d^2(\mathbf{y}_{new},\mathbf{y}_j) \right) \enspace .
\end{split}
\end{equation}

The OOS formula does not depend on the actual value of $\mathbf{y}_{new}$, but rather on the distances between the new response and points on the training dataset. We can thus leverage the predicted $\{\hat{d}(\mathbf{y}_{new},\mathbf{y}_i)\}_{i=1}^N$ on equation \eqref{man:pred:eq:oos} in order to get a point estimate of our response $\hat{\mathbf{z}}_{new}$ on the embedding space $\mathbb{R}^m$. Now we are left with the task of mapping $\hat{\mathbf{z}}_{new}$ to the original manifold ( see figure \ref{man:pred:fig1}(e) ).

\subsubsection{Mapping from the Embedding to the Manifold} \label{sec:back}

The mapping of the response from the embedding to the manifold - backscoring - can be formulated as an interpolation problem. Specifically, we are looking for a smooth continuous function $g: \mathbb{R}^m \rightarrow \mathbb{R}^q $, that minimizes the cost function $\enspace \gamma_g \sum^{N}_{i=1} (g(\mathbf{z}_i)- \mathbf{y}_i)^2  + \|g\|^2_{G} $.
$\gamma_g$ is a weight parameter balancing the smoothness of the interpolation and the adherence to the data and $G$ is a space of smooth functions equipped with the norm $\|\cdot\|_{G}$, which will be constructed in the following.

A solution to the interpolation problem was presented in \cite{wahba1992}. Let $\mathcal{V}$ be a closed subset of $\mathbb{R}^m$ and $k_G$ a kernel function of the form
\begin{equation} \label{man:pred:back:eq2}
k_G(r,\mathbf{v},t,\mathbf{w}), \enspace r,t \in T^q = \{1,\dots, q \}, \mathbf{v}, \mathbf{w} \in \mathcal{V} \enspace .
\end{equation} 
Furthermore, assume that $k_G$ is semi-positive definite on $(T^q \times \mathcal{V}) \times (T^q \times \mathcal{V})$, with 
\begin{equation}
\sum_{i=1}^N \sum_{j=1}^N a_{i} a_{j} \sum_{r=1}^q \sum_{t=1}^q  k_G(r,\mathbf{v}_i,t, \mathbf{v}_j) \geq 0 \enspace,
\end{equation}
for any finite set of points $\{\mathbf{v}_i| \mathbf{v}_i \in \mathcal{V}\}_{i=1}^N$ and real numbers $\{a_i| a_i \in \mathbb{R}\}_{i=1}^N$. For a fixed $(r,\mathbf{v})$, equation \eqref{man:pred:back:eq2} defines a function from $\mathbb{R}^m$ to $\mathbb{R}^q$ by the formula
\begin{equation} \label{man:pred:back:eq4}
g_{r\mathbf{v}} (\mathbf{w}) = \left( k_G(r,\mathbf{v},1,\mathbf{w}), \ldots, k_G(r,\mathbf{v},q,\mathbf{w}) \right)^T \enspace .
\end{equation}

Based on the above, let $G$ be the space of all finite linear combinations of functions of the form \eqref{man:pred:back:eq4}, as $(r,\mathbf{v})$ varies in $T^q \times \mathcal{V}$, and its closure w.r.t the scalar inner product $\langle g_{r\mathbf{v}}, g_{t\mathbf{w}}\rangle = k_G(r,\mathbf{v},t,\mathbf{w})$. It follows that $\| g_{r\mathbf{v}} \|_G = \sqrt{\langle g_{r\mathbf{v}}, g_{r\mathbf{v}} \rangle }$.

The interpolation problem can be solved over the space of functions $G$ as follows \cite{wahba1992}. Let $\mathbf{Y}$ be the $N \times q$ matrix of training responses with rows $Y_{i\cdot} = \mathbf{y}_i$, and $\mathbf{K}^G$ the $Nq \times Nq$ matrix of kernel values $K_{N(r-1)+i,N(t-1)+j}^G = k_G(r,\mathbf{z}_i,t,\mathbf{z}_j)$. The minimizing function can be estimated as:
\begin{equation} \label{man:pred:back:eq5}
\hat{\mathbf{y}}_{new} = \hat{g}(\hat{\mathbf{z}}_{new}) = \sum_{i=1}^{N} \sum_{r=1}^{q} C_{ir} g_{r\mathbf{z}_i}(\hat{\mathbf{z}}_{new})  \enspace ,
\end{equation}
where $C_{ir}$ are elements of the $N \times q$ coefficient matrix $\mathbf{C}$ given by
\begin{equation} \label{man:pred:back:eq6}
vec(\mathbf{C}) = ( \mathbf{K}^{G} + \frac{N}{\gamma_g}\mathbf{I})^{-1} vec(\mathbf{Y}) \enspace ,
\end{equation}
with $vec(\cdot)$ denotes the vectorization of a matrix into a column vector.

It is clear, from equations \eqref{man:pred:back:eq5} and \eqref{man:pred:back:eq6}, that the estimation of $\hat{g}(\hat{\mathbf{z}}_{new})$ requires just the knowledge of pairwise kernel values between the $N+1$ points $\mathbf{z}_1. \ldots, \mathbf{z}_N, \hat{\mathbf{z}}_{new}$.

In our studies, we opted to simplify the minimizer \eqref{man:pred:back:eq5}, in favor of having just one tuning parameter, by choosing
\begin{equation}
k_G(r,\mathbf{v},t,\mathbf{w}) = \begin{cases}
\exp \left( - \frac{\| \mathbf{v} - \mathbf{w} \|^2}{\sigma_G} \right), \enspace r,t = 1, \ldots , q, \enspace r = t\\
0, r \neq t
\end{cases}
\enspace ,
\end{equation}
where $\sigma_G$ is a free parameter adjusting the bandwidth of the kernels.

We notice that the backscoring formulation does not take into explicit consideration the response manifold and $\hat{g}(\hat{\mathbf{z}}_{new})$ is not guaranteed to lie exactly on $\mathcal{M}$. Nevertheless, we justify our choice by pointing that since $\hat{g}$ is a smooth interpolating function from the embedding to the training responses, predictions should also adhere well to the manifold.

\section{Experiments} \label{sec:exp}

In this section, we present a comparative simulation study to assess the ability of our methodology to cast predictions that adhere to the response manifold. Subsequently, we use our method in two illustrative image completion applications, where the objective is to predict one half of an image from the other half.

\subsection{Simulation Study - Prediction on a Swiss-roll Manifold} \label{man:sim}

We simulated $N = 900$ paired input-response points $\{(\mathbf{x}_i, \mathbf{y}_i)\}_{i=1}^N$, with inputs $\mathbf{x}_i \in \mathbb{R}^6$ corresponding to responses $\mathbf{y}_i  \in \mathcal{M}_{sr} \subset \mathbb{R}^3$, where $\mathcal{M}_{sr}$ denotes the 2-dimensional swiss roll manifold embedded in $\mathbb{R}^3$. Only the first two input dimensions were built to be predictive of the output. In detail, we first sampled $900$ points $\{t_i\}_{i=1}^N$ from the uniform distribution $\mathcal{U}(\pi,3\pi)$ and $\{u_i\}_{i=1}^N$ from $\mathcal{U}(0,21)$. The response points on the swiss-roll were computed as $y_{i1} = t_i \cos t_i, \enspace y_{i2} = u_i, \enspace y_{i3} = t_i \sin t_i$, while the input variables $x_{i1}$ and $x_{i2}$ were computed by mapping $u_i$ and $t_i$ in the unit circle: $x_{i1} = \frac{t_i -\bar{t_i}}{\max t_i} \sqrt{1 - \frac{1}{2} \left( \frac{u_i -\bar{u_i}}{\max u_i} \right)^2 }, \enspace x_{i2} = \frac{u_i -\bar{u_i}}{\max u_i} \sqrt{1 - \frac{1}{2} \left( \frac{t_i -\bar{t_i}}{\max t_i}  \right)^2 }$ .
The values for the remaining input coordinate variables were drawn from the standard normal Gaussian distribution. Gaussian Noise, with $\Sigma=0.5\mathbf{I}_3$, was added on the response points.

We compared our dRF prediction methodology to $k$NN regression and standard RF, which do not take into consideration the special form of the response space, as well as kernel RF (kRF) \cite{Geurts2006}, an RF method that employs a kernel function to capture the structure of the response space during training. The simulated dataset was split into $S_{train}$, comprising of $600$ input-response pairs and $S_{test}$, consisting of the remaining $300$. For $k$NN regression, the value of $k$ was selected to be $5$. All RFs were built with 150 trees, no pruning and the number of candidate split features in each node was set to $3$. For kRF, the training gram matrix was calculated using the Gaussian kernel $g(\mathbf{y}_i, \mathbf{y}) = \exp{\left(- \frac{ \|\mathbf{y}_i - \mathbf{y}_j\|^2}{2 \sigma^2} \right) }$ with $\sigma = 2.5$. We followed the prediction methodology as described in \cite{Geurts2006}, with $\mathbf{y}_{new} = \argmin_{\mathbf{y} \in S_{train}} \left( g(\mathbf{y}, \mathbf{y}) -2 \sum_{i=1}^N a(\mathbf{x}_{new}, \mathbf{x}_{i}) g(\mathbf{y}, \mathbf{y}_i) \right)$,
where $a(\cdot,\cdot)$ is the RF-based affinity. The minimization problem was solved over the training set.

For dRF, the backscoring parameters were $\sigma_{\mathcal{G}} = 100$ and $\gamma_{\mathcal{G}} = 200$. Since there is no analytical form for computing distances on a swiss-roll manifold, we approximated manifold distances using the Isomap distance formulation  \cite{Tenenbaum2000}: A neighborhood graph $G$ was constructed, in which each point $\mathbf{y}_i$ was connected to its $k=7$ nearest neighbors in $\mathbb{R}^q$ . Graph edges were assigned weights equal to the Euclidean distances between the connected points in $\mathbb{R}^q$. For any two points $\mathbf{y}_i$ and $\mathbf{y}_j$ in $S$, $d(\mathbf{y}_i,\mathbf{y}_j)$ was then estimated by the shortest path connecting $\mathbf{y}_i$ and $\mathbf{y}_j$ in graph $G$.

\begin{figure}[!ht]
  \centering
  \includegraphics[width=0.6\textwidth]{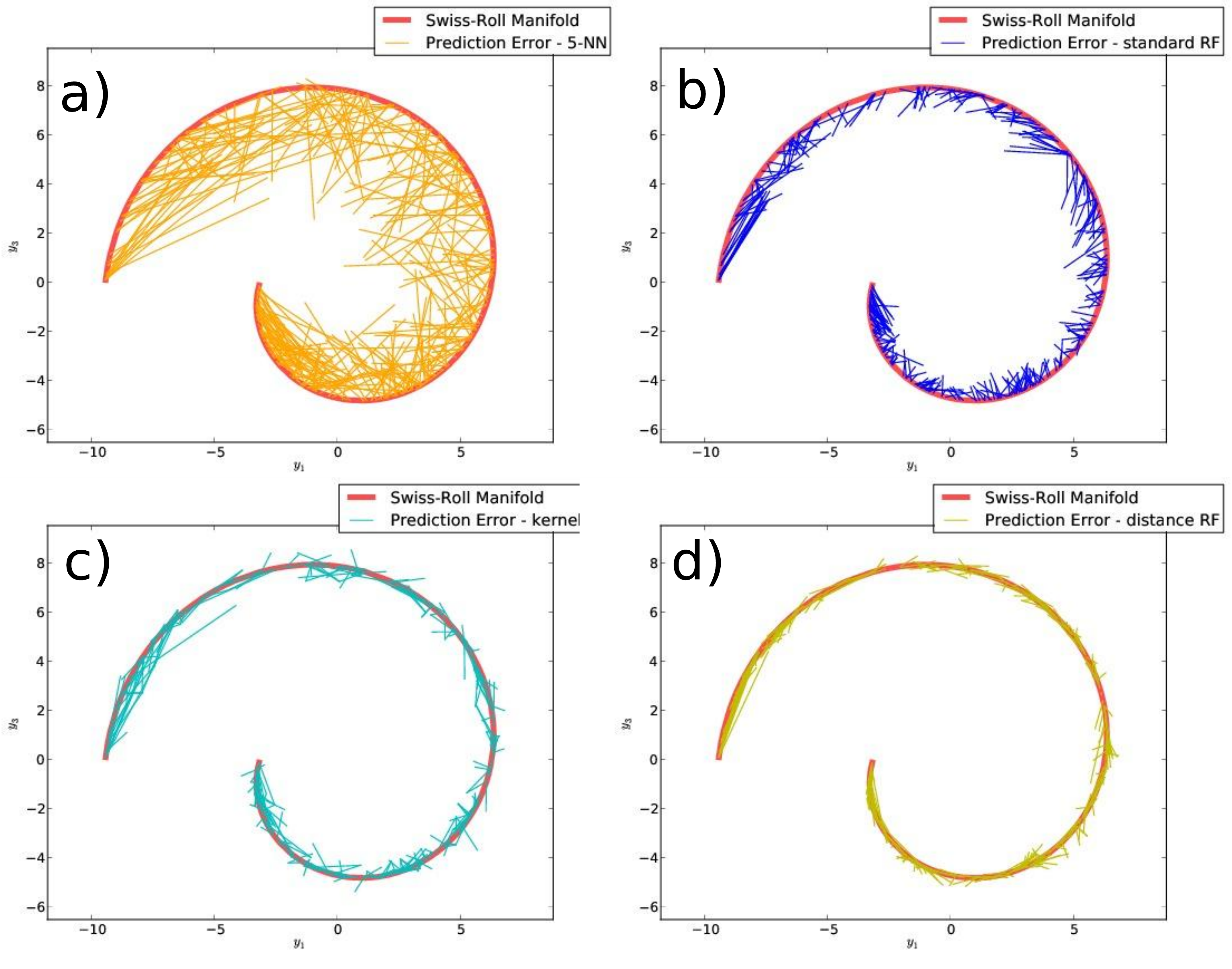}
  \caption {Test error vectors projected on the $y_{\cdot 1}y_{\cdot 3}$ plane of the various regression models for the simulated swiss-roll dataset. (a) $5$NN (b) RF (c) kRF (d) dRF. The figure highlights compliance to the response manifold.}
\label{man:sim:figure9}
\end{figure}

Figure \ref{man:sim:figure9} shows test error vectors for the various methodologies used, projected on the $y_{\cdot 1}y_{\cdot 3}$ plane. We can see that $5$NN missed the goal of regressing on the manifold. For standard RF, it is clear that the model constantly underestimated the radius of the predicted points around the $y_{\cdot 1}$ axis. This can be justified by considering that predictions are taken as average points on the euclidean space $\mathbb{R}^3$, unaware of a possible structure in the response space. kRF preserved the manifold structure of the predicted points better than RF, but suffered significantly from the added noise in the responses. dRF outperformed the other methods both in terms of compliance to the underlying response manifold and regarding good robustness to the addition of noise.

\subsection{Applications on Image Completion} \label{man:app}

In imaging analysis, it is common to assume that a collection of similar images lies on a manifold \cite{peyre2009}. This assumption is guided by the complex nature of images as data objects, as well as the observation that the Euclidean metric and the corresponding geometric structure that it imposes on the space do not bode well with the human perception of similarity and difference between images \cite{peyre2009}. Here, we used our manifold regression methodology to predict the bottom half of handwritten digits and human faces from their upper half.

\subsubsection{Handwritten digits}

\begin{figure}[!ht]
  \centering
  \includegraphics[width=0.4\textwidth]{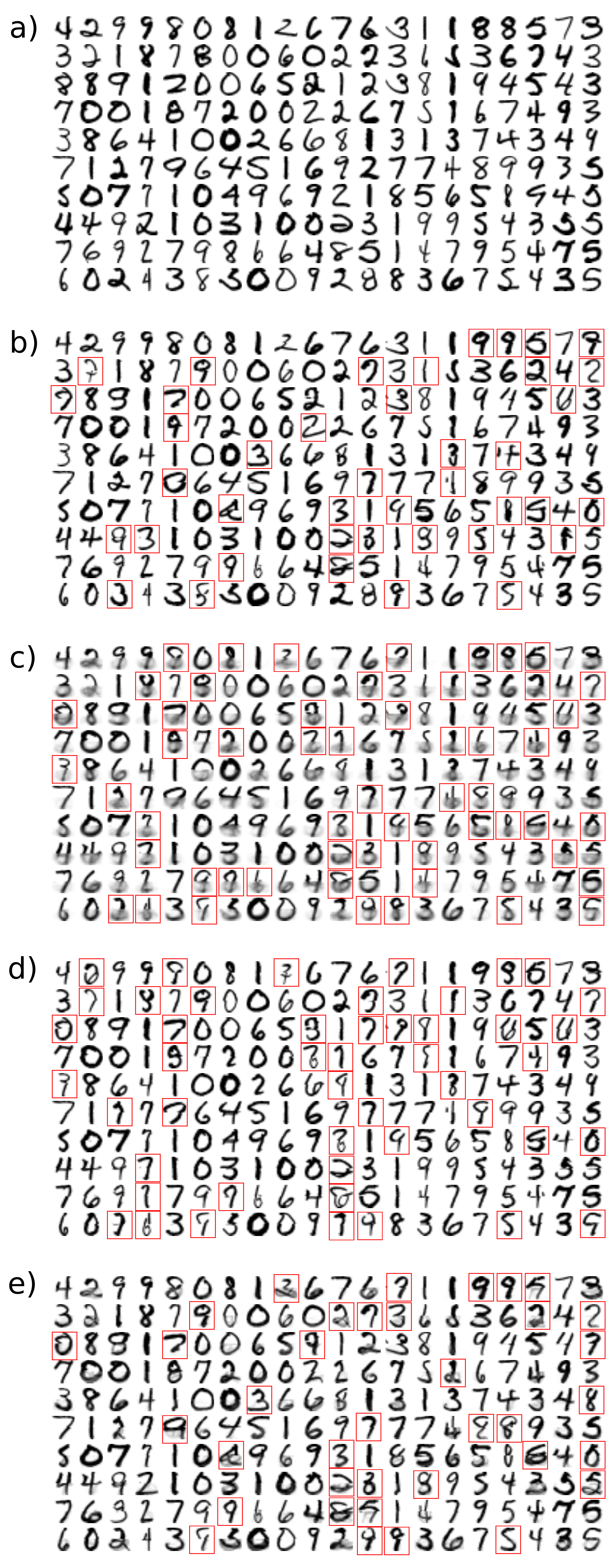}
  \caption{Test and reconstructed images from the digit completion application. Reconstructed digits were generated by concatenating the predicted responses with the upper -input- half of the test images. Bad reconstructions are enclosed in red squares. (a) Test digits. (b)-(e) Reconstructions using predictions from (b) $1$NN, (c) RF, (d) kRF and (e) dRF.}
\label{man:app:figureall}
\end{figure}

For this application, we extracted $1000$ gray-scale images of handwritten digits, from the UCI Machine Learning Repository \cite{Lichman2013}. Each digit class, from $0$ to $9$, was represented in the dataset by $100$ $8 \times 8$ pixel images. Input data were constructed by vectorization of the $8 \times 4$ upper half pixel intensities. The dataset was split into training and testing subsets with $N_{train} = 800$ and  $ N_{test} = 200$. The upper part of each images was taken as input for the predictive models. Responses comprised of the images' bottom parts. The test set can be seen in Fig. \ref{man:app:figureall}(a).

We compared predictions from $1$NN, RF, kRF and dRF models. All RFs were built with $300$ trees and $5$ candidate split features in each node. For dRF, the distance matrix was constructed using the Isomap distance, with the number of neighbors set to $5$. A $25$-dimensional embedding space was used and the backscoring parameters were $\sigma_{\mathcal{G}} = 3$ and $\gamma_{\mathcal{G}} = 20$. For kRF, the training gram matrix was calculated using the Gaussian kernel with $\sigma = 5$. The reconstructed test digits for the various models are shown in Fig. \ref{man:app:figureall}(b)-(e). 

Table \ref{man:app:table1} summarizes the prediction results for the test data. In the first column we include the test Euclidean Mean Squared Errors (EMSE) for all models. In the case of $1$NN and kRF, which draw predictions from the training dataset, we are also able to report misclassification errors, a.k.a the percentage of predicted lower parts that mismatched the ground truth, which are shown on the second column of Table \ref{man:app:table1}. Finally, in the last two columns, we report on the number and percentage of badly reconstructed test images from visual inspection, based on the following criteria: blurriness of the reconstructed image, smooth transition between the upper and bottom image parts, correct digit reconstruction. This qualitative performance measure is important due to the non-Euclidean nature of the data, which, as will be discussed below, makes the EMSE unsuitable for judging the predictive performance.

\begin{table}[!ht]
\caption{Test errors from the digit completion application. Classification Error was not applicable for RF and dRF. The number and percentage of badly reconstructed images was visually ascertained from Fig. \ref{man:app:figureall}.}
\centering
\begin{tabular}{| c | r | r | r | r | r |}
\hline
  & MSE & Clas. Error & Bad Rec. No & Bad Rec. $\%$ \\
\hline
1NN & 3.3665 & 0.165 & 40 & 20 \\
RF & 2.7651 & - & 54 & 27 \\
kRF & 3.2675 & 0.21 & 48 & 24 \\
dRF & 3.3287 & - & 37 & 18.5 \\
\hline
\end{tabular}
\label{man:app:table1}
\end{table}

1NN and kRF cast predictions drawn directly from the training images. As such, no blurriness existed on the reconstructed digits of Fig. \ref{man:app:figureall}(b) and \ref{man:app:figureall}(d). Bad reconstructions were either misclassifications or reconstructions where the transition between the upper (input) and lower (predicted) image parts was not smooth. Surprisingly, $1$NN outperformed kRF in terms of classification error, as well as upon visual inspection of the images.

It is obvious from Fig. \ref{man:app:figureall}(c) that standard RF is ill-suited for the specific application. The RF prediction approach of averaging pixel intensities from various images resulted in a high number of blurry and nonsensical digit reconstructions. It is important to notice that while RF performed the worst, based on visual assessment of the reconstructed images, it had the lowest test EMSE, as shown in table \ref{man:app:table1}. This observation highlights the unsuitability of EMSE as a measure of performance, in the case of manifold-valued responses.

Reconstructed digits from our dRF model are shown in Fig. \ref{man:app:figureall}(e). Our predictions were not drawn directly from the training set. Nevertheless, we notice that the large majority of reconstructions had no fuzziness and the transition from upper to lower parts was smooth. For cases where some blurriness could be noticed in the reconstructions, its effect was significantly less severe than in the RF predictions, resulting in the overall lowest number of badly reconstructed digits, based on visual inspection.

\subsubsection{Faces} \label{man:app2}

In the second application, we used $400$ images of faces from the Olivetti dataset, as included in the scikit-learn python package \cite{scikitlearn}. The dataset consisted of ten gray-scale $64 \times 64$ pixel images for each of 40 distinct subjects, with same subject images taken at different times and with varying pose and facial expressions. Input data were constructed by vectorization of the $64 \times 32$ upper half pixel intensities. The dataset was split into training and testing subsets with $N_{train} = 300$ and $N_{test} = 100$. Images of the same subject were only allocated either to the training or the testing set. Again, the upper parts of the images were taken as inputs, while bottom parts as responses.

We compared results from $1$NN, linear Regression (LR), RF and dRF. RFs were built with $300$ trees and $8$ candidate split features in each node. For dRF, the training distance matrix was constructed using the Isomap metric with the number of neighbors set to $5$, a $25$-dimensional embedding space was used and the backscoring parameters were $\sigma_{\mathcal{G}} = 9$ and $\gamma_{\mathcal{G}} = 50$.

\begin{figure}[!ht]
  \centering
  \includegraphics[width=0.4\textwidth]{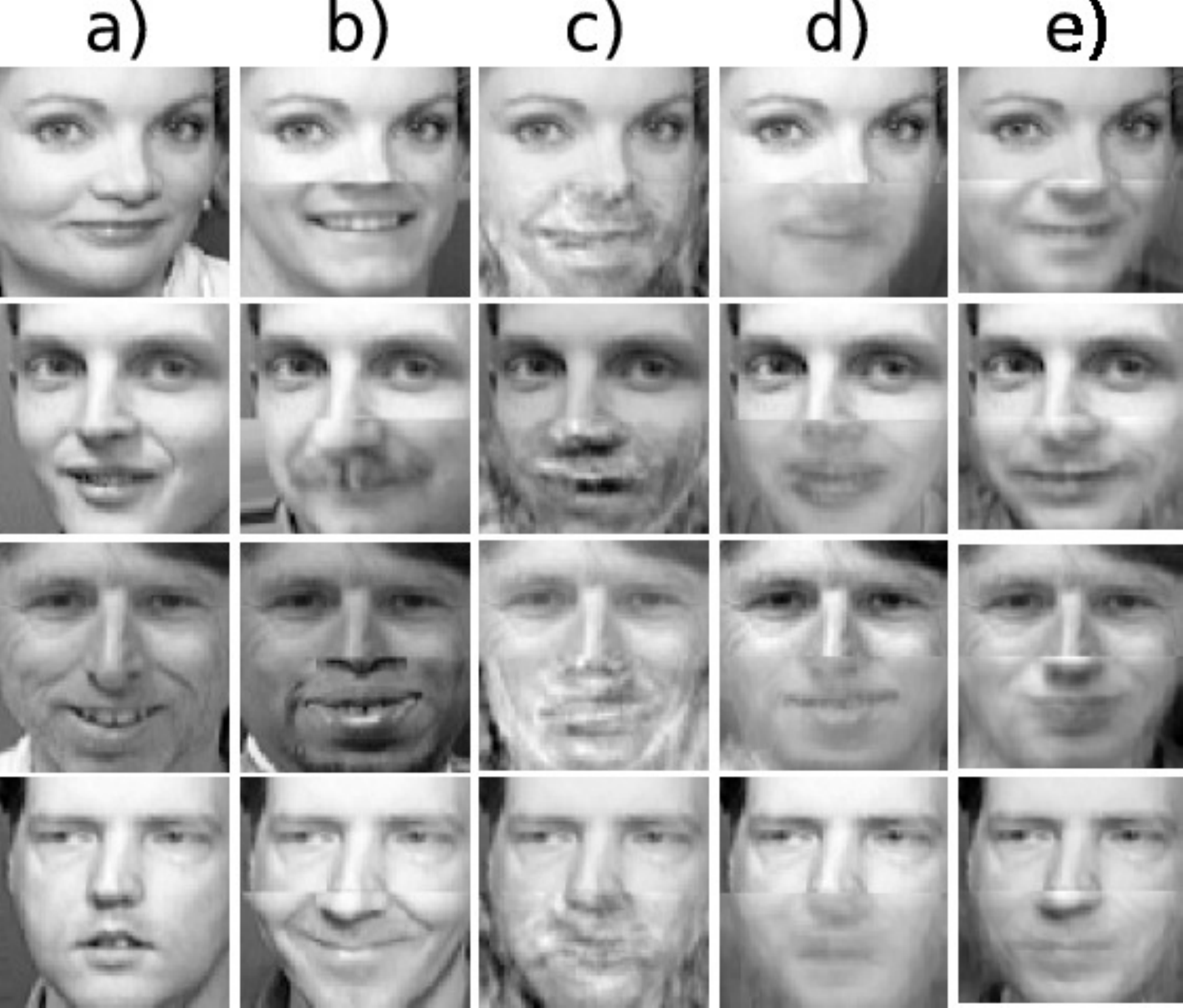}
  \caption{Example test and reconstructed images from the face completion application. (a) Test images. (b)-(e) Reconstructions using predictions from (b) $1$NN, (c) LR, (d) RF and (e) dRF.  }
\label{man:app:figure4}
\end{figure}

Four test images and their reconstructions for the various models are included in Fig. \ref{man:app:figure4}. As we noted in the previous application, EMSE does not provide a suitable descriptor of performance. We rely again on visual inspection of the reconstructed images. 1NN exhibited a hits and miss behavior, with some predictions being similar to the original face, whilst others, such as the second and third depicted faces, being completely different. In addition, there was minimal smoothness in the transitions from the input to the predicted parts, accentuated specifically on the nasal and zygomatic edges. LR predictions were extremely blurry. RF also suffered from a large amount of blurriness, although transitions between the two parts of the faces looked more natural. Finally, dRF reconstructions exhibited the best transition smoothness of all methods, with the predicted half-images being well aligned to their inputs. Although the predictions were not completely free of blur artifacts, the effect was less severe and facial details, such as nasolabial folds and lips, were clearly portrayed.

\section{Discussion and Conclusion} \label{sec:disc}

In this work we presented a predictive modeling approach for response objects occupying non-linear manifold spaces. The regression methodology is based on a distance modification of the RF algorithm that we previously published \cite{Sim2013}, which decouples the model's training from the problem of response representation. For prediction purposes, we constructed a framework in which point estimates are first predicted on a Euclidean embedding of the response manifold, learned from the training dataset, and then projected back on the original space.

Our methodology, in contrast to intrinsic manifold methods, necessitates just the definition of a meaningful distance metric in the response space. This can be especially useful for real-life applications, such as image analysis, where the underlying manifolds are usually not known. Our distance-based regression algorithm draws similarities to the family of kernel-based methodologies. One benefit over kernel methods is the vast library of readily available distance metrics for a plethora of data objects. Furthermore, our methodology presents a unified framework, which deals with backscoring to the original response space, an issue that a lot of presented kernel methods do not tackle sufficiently well.

In the performed experiments, our method showed superior predictive performance in comparison to various regression methods, whilst being able to handle high-dimensional inputs and noise on the response observations. In the future, we aim to investigate the problem of automatic estimation of the Euclidean embedding's dimensionality, as well as the use of more elaborate kernel functions in the backscoring formulation.

\section*{Acknowledgments}
This research did not receive any specific grant from funding agencies in the public, commercial, or not-for-profit sectors. The article is currently under consideration for publication at Patter Recognition Letters.

\end{document}